\documentclass[a4paper, 10pt, conference]{ieeeconf}      
\usepackage{FG2020}

\FGfinalcopy 

\IEEEoverridecommandlockouts                              
\usepackage{graphicx}
\graphicspath{{figures/}}
\usepackage{epsfig} 
\usepackage{mathptmx} 
\usepackage{amsmath} 
\usepackage{amssymb}  
\usepackage{pgfplots}
\usepackage{hyperref}

\def\FGPaperID{4} 

\title{\LARGE \bf
Two-Stream Aural-Visual Affect Analysis in the Wild
}

\author{\parbox{16cm}{\centering
    {\large Felix Kuhnke$^*$, Lars Rumberg$^*$, J{\"o}rn Ostermann}\\
    {\normalsize
    Institut f{\"u}r Informationsverarbeitung, Leibniz University Hannover, Germany\\}}
}

\begin{document}

\ifFGfinal
\thispagestyle{empty}
\pagestyle{empty}
\else
\author{Anonymous FG2020 Workshop submission\\ Paper ID \FGPaperID \\}
\pagestyle{plain}
\fi
\maketitle

\def\thefootnote{*}\footnotetext{These authors contributed equally to this work}\def\thefootnote{\arabic{footnote}}

\begin{abstract}

Human affect recognition is an essential part of natural human-computer interaction. 
However, current methods are still in their infancy, especially for in-the-wild data.
In this work, we introduce our submission to the Affective Behavior Analysis in-the-wild (ABAW) 2020 competition.
We propose a two-stream aural-visual analysis model to recognize affective behavior from videos. 
Audio and image streams are first processed separately and fed into a convolutional neural network. 
Instead of applying recurrent architectures for temporal analysis we only use temporal convolutions. Furthermore, the model is given access to additional features extracted during face-alignment. 
At training time, we exploit correlations between different emotion representations to improve performance.
Our model achieves promising results on the challenging Aff-Wild2 database.

The code is publicly available\footnote{\url{https://github.com/kuhnkeF/ABAW2020TNT}\label{code}}.

\end{abstract}

\section{INTRODUCTION}
Recognizing human affect, the state of emotions and mood, is becoming a crucial part of many human-computer interaction systems. Intelligent systems such as digital or robot assistants for health care applications benefit from the ability to understand human feelings and behaviors. Only if these abilities become mature enough, natural audio-visual human computer interaction will become useful and applicable.

The Affective Behavior Analysis in-the-wild (ABAW) competition by Kollias et al. \cite{kollias2019expression, kollias2018aff, kollias2018multi, kollias2019deep, zafeiriou2017aff,kollias2017recognition, kollias2020ABAWCompetition} tackles this problem by providing a benchmark database and by hosting challenges for three different recognition tasks. 

The benchmark database Aff-Wild2 \cite{kollias2018aff} is composed of in-the-wild videos collected from YouTube. 
The videos include a wide range of content (amateur to professional actors/YouTubers, ethnic groups, ages), styles (editing, shots), situations (hand-held recordings, studio settings, indoor/outdoor) and video qualities (compression settings, video resolutions).
Affect analysis in this in-the-wild setting is much more difficult than for other more regulated datasets.

The challenge tasks for affect analysis include valence-arousal estimation, basic expression recognition and action unit detection (based on the Facial Action Unit Coding System (FACS) \cite{ekman1977facial}). Aff-Wild2 includes annotations for all of these tasks.

We propose to solve the tasks jointly using a multi-task audio-visual recognition model. Multi-task learning of facial behavior and features has been shown to provide better performance than training on a single task \cite{kollias2019face, ranjan2016Hyperface}.
Furthermore, the three different recognition tasks are strongly correlated.
Using the correlation between two of the tasks, by coupling the tasks of action unit detection and categorical emotion classification, has been shown to increase performance \cite{kollias2019face}.
In this work we couple categorical emotion classification with valence-arousal estimation. 
This allows us to create additional pseudo labels that we use during model training.

Our model analyzes the visual and aural video data by using a two-stream approach.
For visual analysis we are inspired by recent works in action recognition. 
We propose to use 3D ResNets \cite{tran2018closer} for affect recognition. 
We divide the video data into short clips that can be processed by spatio-temporal convolutions. 
In addition, we add knowledge from face-alignment to guide our model to learn person-independent face-region-related features.
For aural analysis we implement spatio-temporal convolutions by analyzing audio spectrograms using conventional convolutional neural networks. 
The effectiveness of this approach was investigated 
by Hershey et al. \cite{hershey2017cnn}.
Both modal streams, aural and visual, are merged to produce the final model prediction.

\section{Related Work}

\begin{figure*}[ht]
    \centering
    \includegraphics[width=\textwidth]{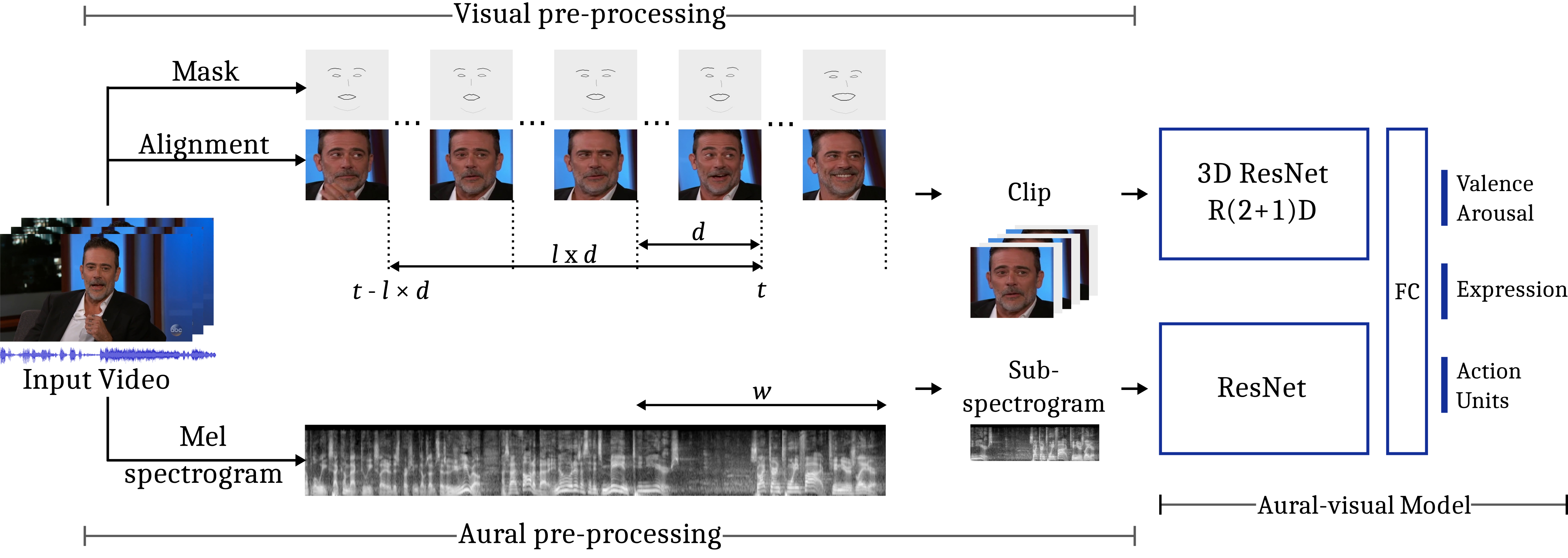}
    \caption{Overview of our method including pre-processing and two-stream aural-visual network. The video is first divided into images and audio to be pre-processed individually. For every frame time $t$ the pre-processed streams provide video clips and sub-spectrograms. These are fed into the corresponding sub-modules of the aural-visual network to produce predictions for three emotion representations.}
    \label{fig:pipeline}
\end{figure*}

Estimating different representations of human emotions or facial manifestations of them, like action units (AU), basic facial expressions (EX) and valance arousal (VA), has been a long-time interest of research. 
We will cover the latest related work briefly.

Recently, multi-task approaches that estimate multiple emotion representations at the same time, have been proposed \cite{kollias2019expression, kollias2019face, chang2017fatauva}. 

Kollias et al. \cite{kollias2019expression} propose a multi-task CNN combined with a recurrent neural network (RNN) for VA and EX recognition. 
They suggest to use ArcFace loss \cite{deng2018arcface} to train a model for expression recognition. 
Furthermore, audio and video data is separately processed by a different CNN, respectively. 
The outputs of both CNNs are processed by a RNN to create the final emotion predictions.
In an additional work, Kollias et al. \cite{kollias2019face} propose to estimate all three representations VA, EX and AU. 
They investigate possible gains from coupling of EX and AU by co-annotation and distribution matching.
Similar to Kollias et al. we investigate the effects of utilizing the correlation between different emotion representations to enrich the training with pseudo labels.
However, instead of EX and AU coupling we investigate EX and VA coupling.

Chang et al. \cite{chang2017fatauva} propose to utilize face attributes as an additional knowledge source for AU and VA estimation.
They follow a hierarchical approach, where face attributes (e.g. gender, age, attractiveness) are learned as outputs of a lower level of a convolutional neural network (CNN). 
The mid level produces AU outputs and the highest level VA estimates.

Lee et al. \cite{lee2019context} propose to use a two-stream 3D-CNN for EX classification. Features are computed separately on the face region and on a context region of the image and then merged by a fusion network. 
Compared to Lee et al., we do not consider a context region, but use the audio signal of a video as our second stream.

\section{METHOD}

\begin{figure*}[thpb]
    \centering
    \scalebox{0.8}{\input{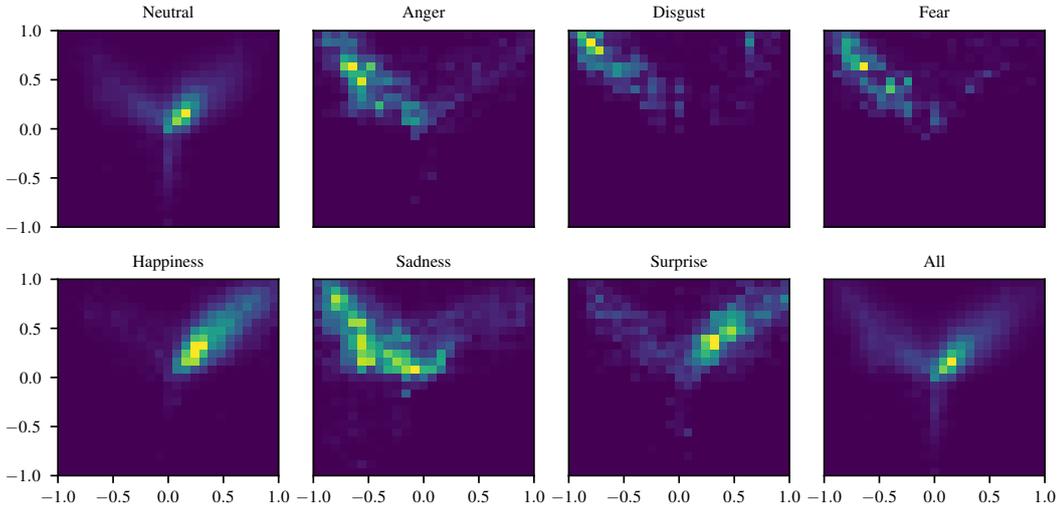}}
        
    \caption{Histograms of valence and arousal for each of the seven basic expressions. Each histogram is individually scaled from zero (dark blue) to the maximum amount of samples in a bin (yellow).}
    \label{fig:ex_va_hist}

\end{figure*}

In this section we introduce our method for affective behavior analysis. 
The overall pipeline of our method is outlined in Fig. \ref{fig:pipeline}. We start by splitting the video into image and audio streams. These streams are pre-processed individually before they are synchronously fed into our two-stream aural-visual model. The model is composed of two sub-modules, each for one stream (aural and visual). At the final model layer both streams are concatenated before joint prediction of three different emotion representations.


\subsection{Visual stream pre-processing}

As a first step, for every video frame we apply the RetinaFace detector \cite{deng2019retinaface}. This provides us with bounding boxes for every face. Aff-Wild2 contains a lot of in-the-wild videos, with multiple faces visible per frame. We use ArcFace \cite{deng2018arcface} and heuristics like bounding box size to annotate the main person in every video. Furthermore, we also did some manual cleaning on the results. 
The cropped images from these boxes are fed into PRNet \cite{feng2018prnet} to obtain 68 landmarks for every face detection.
Using a 5 point template (eye centers, nose tip, outer mouth corners) and the corresponding points on the image (based on the landmarks) we perform face alignment using a similarity transform. 
The aligned image is scaled to $height \times width$ pixels.
Additionally, for every aligned face image we render a mask image of the outer eye contours, the nose, the chin, the brows, and the outer lip contour, based on the 68 landmarks. Our face-alignment, landmarks and code for mask rendering is available for further research\footnotemark[1].
Exemplary results for alignment and mask images can be seen in Fig. \ref{fig:pipeline}.

The input to the visual sub-module of our network is a 4 dimensional tensor. 
This tensor represents a short video clip. 
A clip contains $l$ frames, sampled with dilation $d$.
Dilation defines the sampling resolution that is how many frames are skipped between sampling frames.
A clip sampled at time $t$, contains the frame at $t$ and $l-1$ preceding frames. 
As all videos are sampled at 30 fps, the final span of time a clip contains is $l \times d \times \frac{1}{30}$ seconds. 
Every clip frame is a 4 channel image, that is composed of the aligned face image and the mask image. 
Both images are resized to the same width and height. 
The final size of the clip's tensor is $l \times height \times width \times 4$.

\subsection{Aural stream pre-processing}

For aural pre-processing, the audio stream is extracted from the video and resampled to 41 kHz. Afterwards, we compute a mel spectrogramm using the TorchAudio package, which is part of PyTorch \cite{paszke2017pytorch}. We use following settings:
\begin{itemize}
    \item number of mel filter banks $n_\mathrm{mels} = 64$
    \item window size $w_\mathrm{win}  = 20\,$ms
    \item window stride $t_\mathrm{stride}  = 10\,$ms
    \item FFT size $n_\mathrm{FFT}  = 1024\,$bins
\end{itemize}

Before feeding it into the aural sub-module, the spectrogram is cut into a smaller sub-spectrogram of length $w$, with the center of sub-spectrogram aligning with the current frame at time $t$.  
The sub-spectrogram is an image of size $(w \times \frac{1}{t_\mathrm{stride}\,[s]} + 1)  \times n_\mathrm{mels}$. 
For a sub-spectrogram with length of $10\,$s and the settings given above, this results in a tensor of size $1001 \times 64$.

\subsection{Two-Stream Aural-Visual Model}

The model is composed of two sub-modules and one fusion layer. The sub-modules provide aural and visual analysis, respectively. The fusion layer concatenates the output of both sub-models and predicts the three emotion representations using a single fully connected layer. 

The \textbf{visual model} is a variant of 3D ResNets called R(2+1)D proposed by Tran et al. \cite{tran2018closer}. It has performed favorably for action recognition and needs less parameters compared to classical 3D CNNs. The model is pre-trained on the Kinetics-400 database \cite{kay2017kinetics}. As the input clip has 4 channels, we modify the first convolution layer to accept 4 instead of 3 channels. Furthermore, we remove the last fully connected layer.

The \textbf{aural model} analyses the sub-spectrograms.
We use Resnet18 \cite{he2016} to extract features.
Again, the input size channels do not match the original implementation. 
Therefore, we modify the first convolution layer to accept 1 instead of 3 channels. 
Furthermore, we remove the last fully connected layer.

Finally, the output of both sub-models is concatenated and feed into a single fully connected layer. This layer produces the three emotion representations: Continuous valence and arousal, visible basic expression and visible action units.

\subsection{Loss functions}

We use the same loss functions as they are defined in \cite{kollias2019expression}.
The categorical cross entropy for categorical expression classification.
The binary cross entropy for action unit detection and the concordance correlation coefficient loss for valence and arousal estimation. 
We divide each loss by the amount of labeled samples in the current mini-batch and use the sum as training objective.

\section{Data processing}

\subsection{Pseudo labels}
\label{sec:pseudo_labels}


In the training set of the Aff-Wild2 database, only $59\,$\% of all
frames are labeled for categorical expressions and $75\,$\%
are labeled for valence and arousal. 
We use the correlation between the two different emotion representations to enrich the training with pseudo labels.

To achieve this, we use the histograms shown in Fig. \ref{fig:ex_va_hist} to compute pseudo labels for the parts of the database which are only labeled for one of the tasks. 
The histograms show the distribution of valence and arousal labels for each of the categorical expressions, computed over all frames in the training set of the database which are labeled for both categorical (EX) and continuous emotion (VA).


Given an expression label, we sample a valence and arousal label from the distribution of this expression. 
This sampling is redone every time the network processes the frame during training.

Given a valence and arousal label $v$ and $a$ we compute a probability $p_{i}(v,a)$ for each expression $i$ with
\begin{equation}
    p_{i}(v,a) = \frac{n_{i}(v,a)}{\sum\limits_{i \in E}^{} n_{i}(v,a)},
\end{equation}
where $n_{i}(v,a)$ is the amount of valence and arousal labels in the corresponding bin of the histogram for expression $i$ and $E$ is the set of all expressions. These probabilities are used as soft expression labels during training.

This method allows us to create additional labels for the previously unlabeled $25\,$\%  and $31\,$\% of the frames in the training set for VA and EX, respectively.

\begin{table*}[tb]
\caption{
Results on Aff-Wild2 test set 
}
\begin{center}
\def\arraystretch{1.1}
\begin{tabular}{@{\extracolsep{4pt}}lrrrrrrrrr@{}}
\hline
                                                        & \multicolumn{3}{c}{CCC}                       & \multicolumn{3}{c}{Expression}    & \multicolumn{3}{c}{Action Unit} \\
                                                        \cline{2-4}                                     \cline{5-7}                         \cline{8-10}
Method                                                  & Valence       & Arousal       & Mean          & F$_1$    & Acc   & Criterion         & F$_1$    & Acc   & Criterion\\
\hline\hline
Baselines from \cite{kollias2020ABAWCompetition}:       &               &               &               &       &       &                   &       &       & \\
PatchGAN                                                & 0.11          & 0.27          &  0.19         & -     & -     & -                 & -     & -     & -\\
MobileNetV2                                             & -             & -             & -             & -     & -     & 0.30              & -     & -     & -\\
MobileNetV2                                             & -             & -             & -             & -     & -     & -                 & -     & -     & 0.26\\
\hline
Prior work on Aff-Wild2:\\
A/V-MT-VGG-RNN \cite{kollias2019expression}             & 0.38          & 0.42          & 0.4           & 0.46  & -     & -                 & -     & -     & -\\
MT-VGG-RNN \cite{kollias2019expression}                 & -             & -             & -             & -     & -     & -                 & 0.44  & -     & -\\
FaceBehaviorNet no coupling \cite{kollias2019face}      & 0.45          & 0.40          & 0.43          & 0.48  & -     & -                 & 0.47  & -     & -\\
FaceBehaviorNet with coupling \cite{kollias2019face}    & \textbf{0.51}          & \textbf{0.46}          & \textbf{0.49}          & \textbf{0.53}  & -     & -                 & \textbf{0.52}  & -     & -\\
\hline
Top entries to ABAW \cite{kollias2020ABAWCompetition}: \\
NISL2020 \cite{deng2020fau}                             & 0.44          & \textbf{0.45}          & \textbf{0.45}          & 0.27  & 0.68  & 0.41              & \textbf{0.31}  & 0.91  &  \textbf{0.61}\\
ICT-VIPL \cite{zhang2020m}                              & 0.36          & 0.40          & 0.38          & 0.29  & 0.66  & 0.41              & -              & -      &  -\\
SALT \cite{pahl2020SALT}                                & -             & -             & -             & -     & -     & -                 & 0.216          & 0.886  &  0.551\\
TSAV (ours)                                             & \textbf{0.45}          & 0.41          & 0.43          & \textbf{0.40}  & \textbf{0.70}  & \textbf{0.50}     & 0.27           & \textbf{0.93}  & 0.60 \\
\hline

\end{tabular}
\end{center}

\label{tab:results}
\end{table*}

\subsection{Filtering}
\label{sec:filtering}
During analysis of the interplay of different emotion representations, we found that emotion representation annotations (AU, VA, EX) for the same frame can be contradictory. 
E.g. a frame labeled with the "happy" expression can simultaneously be annotated with a negative valence. Such contradictory annotations might hint to either annotator disagreement or faulty annotations.
Furthermore, some labels are not in the defined range, e.g. $[-1,1]$ for VA.  
We implement an effective filtering strategy to exclude such examples from training.
If one of the following conditions is true, the frame is not used during training:
\begin{itemize}
    \item Either the valence and arousal or the expression label is not valid.
    \item The expression is labeled as happy, but the valence is labeled as negative.
    \item The expression is labeled as sad, but the valence is labeled as positive.
    \item The expression is labeled as neutral, but $\sqrt{valence^2 + arousal^2} > 0.5$
\end{itemize}

\section{EXPERIMENTS}

\subsection{Implementation and Setup}

Our framework is implemented using PyTorch \cite{paszke2017pytorch}.
We train our model on Aff-Wild2 with the following parameters:\\
Clip length $l=8$, dilation $d=6$, sub-spectrogram length $w=10$ seconds.
Clip $width$ and $height$ are set to $112$ pixel.

Data augmentations for video clips are random horizontal flip, and small random changes to hue, saturation and lightness.
We use pre-trained networks provided by \cite{paszke2017pytorch}.
The mini-batch size is set to 320. More specifically, this means 320 clips and 320 sub-spectrograms are processed in each iteration.
To be able to compute on one GPU we divide the mini-batch into 10 parts and accumulate the gradients.
Training is done using Adam optimizer \cite{kingma2014adam} and a learning rate of 0.001.

A first model is trained on the original training set of Aff-Wild2. 
We stop the training at the best validation performance and note the number of iterations.
Our best submission is trained on the original validation set and the training set of Aff-Wild2.
Instead of stopping on the best validation performance, we stop at 1.25 times the number of iterations of the previously (training set only) trained model.

Pseudo labels can be created for expression, valence and arousal. However, we only use pseudo labels for valence. 
We found that arousal and expression pseudo labels deteriorate the performance (Section \ref{sec:ablations}).  
Furthermore, we filter the training data set as described in Section \ref{sec:filtering} to reduce the influence of noisy labels during training. 
We remove $6.7\,\%$ of training data:
\begin{itemize}
    \item 6000 frames labeled "happy" but with negative valence.
    \item 13000 frames labeled "sad" but with positive valence.
    \item 121000 frames labeled "neutral" but with high valence arousal norm.
\end{itemize}

\subsection{Results} 
\begin{table*}[tb]
\caption{
Ablation study on Aff-Wild2 validation set.
}
\begin{center}
\def\arraystretch{1.1}
\begin{tabular}{lrrrrrr}
\hline
Method                              & CCC Valence       & CCC Arousal       & CCC Mean          & Expression Criterion  & Action Unit Criterion     & Mean Score \\
\hline\hline
TSAV (Filter \checkmark, pseudo V)  & 0.493             & \textbf{0.613}    & \textbf{0.553}    & \textbf{0.546}        & 0.587                     & \textbf{0.561}\\
\hline
No mask                             & 0.464             & 0.558             & 0.511             & 0.483                 & 0.604                     & 0.532 \\
\cite{kollias2020ABAWCompetition} Alignment, no mask
                                    & 0.493             & 0.608             & 0.551             & 0.478                 & 0.583                     & 0.537\\
\hline
Only visual stream                  & 0.463             & 0.570             & 0.517             & 0.515                 & 0.576                     & 0.536 \\
Only aural stream                   & 0.355             & 0.359             & 0.357             & 0.434                 & 0.468                     & 0.420\\
\hline
Filter \checkmark, pseudo V, A, EX  & 0.496             & 0.543             & 0.519             & 0.416                 & 0.587                     & 0.508 \\
Filter $\times$, pseudo V           & \textbf{0.497}    & 0.586             & 0.541             & 0.507                 & \textbf{0.605}            & 0.551 \\
Filter \checkmark, pseudo $\times$  & 0.394             & 0.591             & 0.492             & 0.531                 & 0.587                     & 0.537 \\
Filter $\times$, pseudo $\times$    & 0.404             & 0.581             & 0.493             & 0.508                 & 0.588                     & 0.530 \\
\hline

\end{tabular}
\end{center}

\label{tab:ablations}
\end{table*}
Table \ref{tab:results} shows our results on the test set of Aff-Wild2. We report the same evaluation criterion as outlined in \cite{kollias2020ABAWCompetition}. Valence-Arousal estimation is based on the mean Concordance Correlation Coefficient (CCC). Classification of the seven basic expressions is measured by $0.67 \times\, \mathrm{F_1\:score} + 0.33 \times\, \mathrm{total\:accuracy}$ (Expression Criterion). The overall score for AU detection is $0.5 \times\, \mathrm{average\:F_1\:score} + 0.5 \times\, \mathrm{total\:accuracy}$ (Action Unit Criterion). 

Our model significantly outperforms the baselines provided in \cite{kollias2020ABAWCompetition}. In contrast to our work, the baseline models are single-task networks which analyze only single images.
In the following we will compare out results to prior works on Aff-Wild2 by Kollias et al. and to the top three entries of every recognition task (VA, EX, AU) of the ABAW 2020 competition.

\textbf{VA}: 
\cite{kollias2019expression} is a multi-task network incorporating aural and temporal information. Our model has better performance for valence and only slightly worse performance for arousal. \cite{kollias2019expression} uses a RNN to incorporate temporal information, however, we use 3D convolutions.
Similar to our approach, the models of \cite{kollias2019face} and Deng et al. \cite{deng2020fau} both use inter-task correlations. 
Compared to our model, both have slightly better performance  for valence and arousal estimation. However, both use additional datasets for training, whereas we only use the Aff-Wild2 database. 

\textbf{EX}: 
For expression recognition, our method outperforms all ABAW competitors.
One reason for this could be the use of mask images, which gives the network a prior on face appearance and position. 
This is also suggested by our ablation study in Section \ref{sec:ablations}.
Surprisingly, the $\mathrm{F_1\:score}$ of \cite{kollias2019expression} and \cite{kollias2019face} is better than ours.
In interpreting these results, we have to take into account that FaceBehaviorNet \cite{kollias2019face} is trained on several facial affect databases, whereas our model is only trained on Aff-Wild2. 
Conversely, MT-VGG-RNN \cite{kollias2019expression} was only trained on Aff-Wild2. 
Future research should attempt to clarify these differences.

\textbf{AU}:
\cite{deng2020fau} surpass our results for AU recognition by a small margin. Notably, our $\mathrm{F_1\:score}$ is lower than \cite{deng2020fau}'s, but our accuracy is higher.
We assume this is due to their data balancing strategy, which increases their $\mathrm{F}_{1}$ score on rare AUs but is detrimental to learning the dataset overall statistics which leads to a lower accuracy. Again, their model uses additional datasets for training.

Interestingly, \cite{kollias2019expression, kollias2019face} surpass all competition results for AU $\mathrm{F_1\:score}$.
Similar to EX performance, a more detailed analysis of methods and results is necessary to compare the methods properly.


\subsection{Ablations}
\label{sec:ablations}
To analyze the effects of different alignments, mask input, network structure and data processing we conducted additional experiments.
The results can be seen in Table \ref{tab:ablations}. 

The usage of mask images as an additional input to the model leads to significantly increased performance of EX classification.
The mask images most likely help our model to generalize to unseen faces and provide a strong prior during training. 

While using pseudo labels for valence increases the performance of the model, using pseudo-labels for EX recognition significantly decreases accuracy. 
Further investigations show, that the pseudo-labels for EX are especially detrimental to the performance for the expressions which are rare in the training data and have a high overlap with a more common expression in the valence-arousal space.
This overlap leads to pseudo-labels which are biased towards the common expressions.
Interestingly, filtering the data only improves expression recognition. 

As expected, combining both the aural and the visual stream increases performance for all tasks, especially VA estimation and EX classification.


\section{CONCLUSIONS}

We proposed a new fully convolutional model for audio-visual affect recognition.
Our results on the challenging Aff-Wild2 database are promising and outperform the baseline by a wide margin.
By rendering masks from features extracted during face-alignment and using them as additional information during training, our model is able to outperform all competing models in the ABAW 2020 competition for classification of basic expressions.
Our model also shows strong results for action unit and valence arousal estimation, although it is only trained on one data set.

In future work we will investigate data balancing strategies and additional methods for fusing the aural and visual streams.




\bibliographystyle{ieee.bst}
\bibliography{affwild.bib}

\end{document}